\documentclass[mlmain]{jmlr}

\usepackage{longtable}

\usepackage{xcolor}       

\usepackage{bm}
\usepackage{algorithm}
\usepackage{algorithmic}
\usepackage{multirow}
\usepackage{svg}

\usepackage{booktabs}
\usepackage[load-configurations=version-1]{siunitx} 

\def \real{\mathbb{R}}
\def \x{\mathbf{x}}
\def \y{\mathbf{y}}

\def \dec{\mathsf{cl}}
\def \out{\mathcal{O}}

\def \dout{\partial \out}

\def \dim{D}

\theorembodyfont{\upshape}
\theoremheaderfont{\scshape}
\theorempostheader{:}
\theoremsep{\newline}

\jmlrvolume{}
\firstpageno{1}
\editors{List of editors' names}

\jmlryear{2025}
\jmlrworkshop{Symmetry and Geometry in Neural Representations}

\title[Curvature Dynamic Black-box Attack]{Curvature Dynamic Black-box Attack: revisiting adversarial robustness via dynamic curvature estimation}

 \author{\Name{Peiran Sun} \Email{peiransun@qq.com}\\
 \addr Lanzhou University, Lanzhou, China}

\begin{document}

\maketitle

\begin{abstract}
Adversarial attack reveals the vulnerability of deep learning models. It is assumed that high curvature may give rise to rough decision boundary and thus result in less robust models. However, the most commonly used \textit{curvature} is the curvature of loss function, scores or other parameters from within the model as opposed to decision boundary curvature, since the former can be relatively easily formed using second order derivative. In this paper, we propose a new query-efficient method, dynamic curvature estimation (DCE), to estimate the decision boundary curvature in a black-box setting. Our approach is based on CGBA, a black-box adversarial attack. By performing DCE on a wide range of classifiers, we discovered, statistically, a connection between decision boundary curvature and adversarial robustness. We also propose a new attack method, curvature dynamic black-box attack (CDBA) with improved performance using the estimated curvature. 
\end{abstract}

\section{Introduction}

Adversarial attacks are generally classified into two types:
white-box \cite{43405,moosavi2016deepfool,7958570} and black-box \cite{chen2017zoo,papernot2016transferability,brendel2017decision,maho2021surfree,rahmati2020geoda,reza2023cgba,wang2022triangle,chen2020hopskipjumpattack} attacks. In white-box setting the
attacker has full knowledge about the interior of the model,
where using gradient descent via backpropagation is a common practice. However, white-box scenario is too ideal for real-world application. For black-
box setting, there are score-based \cite{chen2017zoo}, transfer-based \cite{papernot2016transferability} and decision-based attacks \cite{brendel2017decision,rahmati2020geoda,maho2021surfree,reza2023cgba,chen2020hopskipjumpattack}. Score-based attack means that the attacker can
access the output probability of the last layer. In transfer-based
attack, a surrogate model is trained to generate adversarial
examples. The decision-based attack is the most practical, yet hardest attack since it only queries the model for the top-1 label.

Early analysis \cite{7958570} of adversarial robustness tried to interpret robustness as resilience to attacks. Specifically, there are $\ell_p$ robustness \cite{croce2021robustbench} and certified robustness \cite{pmlr-v97-cohen19c}. These approaches can be effective in adversarial training, but cannot help to explain the vulnerability of neural networks. \cite{10.5555/3157096.3157279} was among the first researching the relationship between decision boundary curvature and robustness. Instead of studying the quantitative curvature of decision boundaries, it resorted to extensive theoretical proof to show the impact of curvature bound on the relative robustness across different noise regimes. In \cite{8953595}, curvature of the loss function is calculated by the spectral norm of Hessian matrices. In \cite{10.5555/3692070.3692569}, Lipschitz constant was shown to be tight upper bound of the second derivative of the output scores. Curvature regularization via Hessian or Lipschitz constant were proposed to improve robustness.
However, these curvature are \textbf{parameter curvature} as they can be formulated via backpropagation. \textbf{Decision boundary curvature} is hard to be formulated since there's no simple expression for decision boundary.

In this paper, we try to estimate decision boundary curvature of image classifiers in a query-efficient way. We focus on the decision-based adversarial attack and compare robustness with decision boundary curvature. Our contributions are:
\begin{itemize}
    \item We propose a curvature dynamic trajectory alongside the black-box adversarial attack process to perform Dynamic Curvature Estimation (DCE) for every iteration of the attack in the black-box setting.
    \item We find a connection between curvature and adversarial robustness by conducting DCE over the standard and robust models on CIFAR-10 and ImageNet, the curvature results are compared with common robustness metrics, including robust accuracy from Robustbench \cite{croce2021robustbench} and certified accuracy from \cite{pmlr-v97-cohen19c}.
    \item We introduce Curvature Dynamic Black-box Attack (CDBA), by integrating DCE in current state-of-the-art CGBA, to utilize curvature information in adversarial attacks.
\end{itemize}

\section{Related Work}

\subsection{Adversarial Attacks}

Decision-based adversarial attack was first studied in \cite{brendel2017decision}. HSJA \cite{chen2020hopskipjumpattack} introduced normal vector estimation to push the adversarial examples into the vast adversarial space indicated by the normal vector. GeoDA \cite{rahmati2020geoda} was among the first to hypothesize that the decision boundary in non-targeted attacks can approximated by hyperplanes. This hypothesis is the key to the formulation of the semicircular path used in SurFree \cite{maho2021surfree} and CGBA \cite{reza2023cgba} since the closest point on the plane should always lie in the perpendicular direction of that plane, and the angle inscribed across a circle's diameter is always a right angle. The difference between SurFree and CGBA is the choice of the restricted 2D plane on which the semicircular search is performed. In CGBA, the plane is spanned by the adversarial examples and the normal vector of the decision boundary while SurFree employs random orthogonal directions. 

\subsection{Adversarial robustness}

Adversarial robustness is commonly evaluated using adversarial examples. There are $\ell_p$ robustness and certified robustness. In \cite{croce2021robustbench}, $\ell_p$ robustness is evaluated by the robust accuracy of adversarial examples generated by AutoAttack, an ensemble of white- and black-box attacks. In \cite{pmlr-v97-cohen19c}, certified robustness is evaluated by certified accuracy of images sampled near source images with random Gaussian corruptions. 

Common corruption robustness \cite{DBLP:journals/corr/abs-1903-12261} is another realm of robustness since it is evaluated using algorithmically generated images instead of adversarial examples. The key difference is that adversarial robustness always requires a limited perturbation budget. Common corruption robustness, on the other hand, is generally considered to be more practical for real-world applications.

\section{Problem statement}

We introduce the following notations.

The image classifier to be attacked can be written as $f(\x):\real^\dim\to\real^C$, $\x\in\real^\dim$ being the input image, $D$ being the dimension of the input image and $C$ being the number of classes. In the decision-based setting, the only information available for attack is the top-1 label $\dec(\x):=\arg\max_k f_k(\x)$, $f_k(\x)$ standing for the probability predicted for each class $k$, $1\leq k\leq C$.

In non-targeted attacks, the attacker only has one correctly classified image $\x_s$. The goal of the attack is to find a perturbed image $\x_b$
satisfying $\dec(\x_b)\neq\dec(\x_s)$. The adversarial space is defined
by $\out_{\text{non-targeted}} := \{\x_b\in\real^\dim: \dec(\x_b)\neq\dec(\x_s)\}$. The decision boundary is $\dout_{\text{non-targeted}} := \{\x_b\in\real^\dim: f_{\dec(\x_s)}(\x_b)= f_{i}(\x_b), i\neq\dec(\x_s)\}$.

In targeted attacks, the attacker has a correctly classified source image $\x_s$ and a target image $\x_t$. The goal of the attack is to find a perturbed image $\x_b$
satisfying $\dec(\x_b)=\dec(\x_t)$. The adversarial space is defined
by $\out_{\text{targeted}} := \{\x_b\in\real^\dim: \dec(\x_b) = \dec(\x_t)\}$.
 The decision boundary is $\dout_{\text{targeted}} := \{\x_b\in\real^\dim: f_{\dec(\x_s)}(\x_b)= f_{\dec(\x_t)}(\x_b)\}$.

We simplify the notation for two adversarial spaces $\out_{\text{non-targeted}}$, $\out_{\text{targeted}}$ as $\out$ where it's applicable for both. The optimization of adversarial attacks can be written as:

\begin{equation}
\label{eq:OptDef}
\x_b^\star = \arg\min_{\x\in\out} \|\x_b-\x_s\|.
\end{equation}

\section{Proposed Methods}

We carry out the attack following CGBA \cite{reza2023cgba}, by conducting boundary search on normal vector guided semicircular path. Specifically, for a clean image $\x_s$ and adversarial point $\x_{b_t}\in\dout$ at $t$th iteration, we first perform normal vector estimation of the decision boundary as \cite{rahmati2020geoda}:
\begin{equation} 
    \hat{\bm \eta}_t = \frac{\sum_{i=1}^{N_t} \phi(\x_{b_t} + \bm\delta_i)\bm \delta_i}{\|\sum_{i=1}^{N_t} \phi(\x_{b_t} + \bm \delta_i)\bm \delta_i\|_2} .
\end{equation}
where $\bm \delta_i \sim \mathcal N(0, \mathbf{\sigma^2})$, $N_t$ is the number of queries used for estimation and $\phi(\cdot)$ is the hard label indicator function:
\begin{equation}
    \;\; \phi(\x)= 
    \begin{cases}
    1, & \text{if  }\; \x\in\out\\
    -1, & \text{otherwise}
    \end{cases}
\end{equation}

Since the semicircular search and DCE are all done on the restricted plane spanned by $\x_{b_t} - \x_s $ and the estimated normal vector $\hat{\bm \eta}_t$, we wish to introduce coordinates on the plane for the sake of discussion. We set the origin at $\x_s$, set the $\frac{\x_{b_t} - \x_s}{\| \bm x_{b_t} - \bm x_s\|_2}$ as the x-axis base $\x_i$, use Schmidt orthogonalization on $\hat{\bm \eta}_t$ to get the normalized perpendicular direction for y-axis base $\y_i$. 

It should be noted that there is no guarantee that the estimated  $\hat{\bm \eta}_t$ is perpendicular to the decision boundary on this plane since $N_t$ is far less than $D$ (this will be further discussed in ablation study). But in general, the orthogonal $\y_i$ still points towards the adversarial region.

Then, any point on the spanned plane can be written as:
\begin{equation}
\x = \x_s + \langle\x-\x_s,\x_i\rangle\x_i + \langle\x-\x_s,\y_i\rangle\y_i
\end{equation}

where $\langle ,\rangle$ denotes the inner product. For simplicity, we denote the coordinates $(\langle\x-\x_s,\x_i\rangle,\langle\x-\x_s,\y_i\rangle)$ as $(x,y)$. Similarly we can get the polar coordinates $(\rho,\theta)$ by $(\frac{\|\x - \x_s\|_2}{\| \bm x_{b_t} - \bm x_s\|_2},\arctan\frac{\langle\x-\x_s,\y_i\rangle}{\langle\x-\x_s,\x_i\rangle})$.

The semicircular path used in CGBA can be written in polar coordinates as:
\begin{equation}
    \rho=cos(\theta)\;\;\; \text{s.t.}\;\theta\in[0,\pi/2]
\label{eq:circle}
\end{equation}
By binary searching on $\theta$ until the error between adversarial and clean points is less than a set tolerance, the algorithm determines the boundary point $\x_{b_{t+1}}\in\dout$ for next iteration.

\subsection{Dynamic Curvature Estimation (DCE)}
\begin{figure}[htbp]
\centerline{\includegraphics[width=0.6\textwidth]{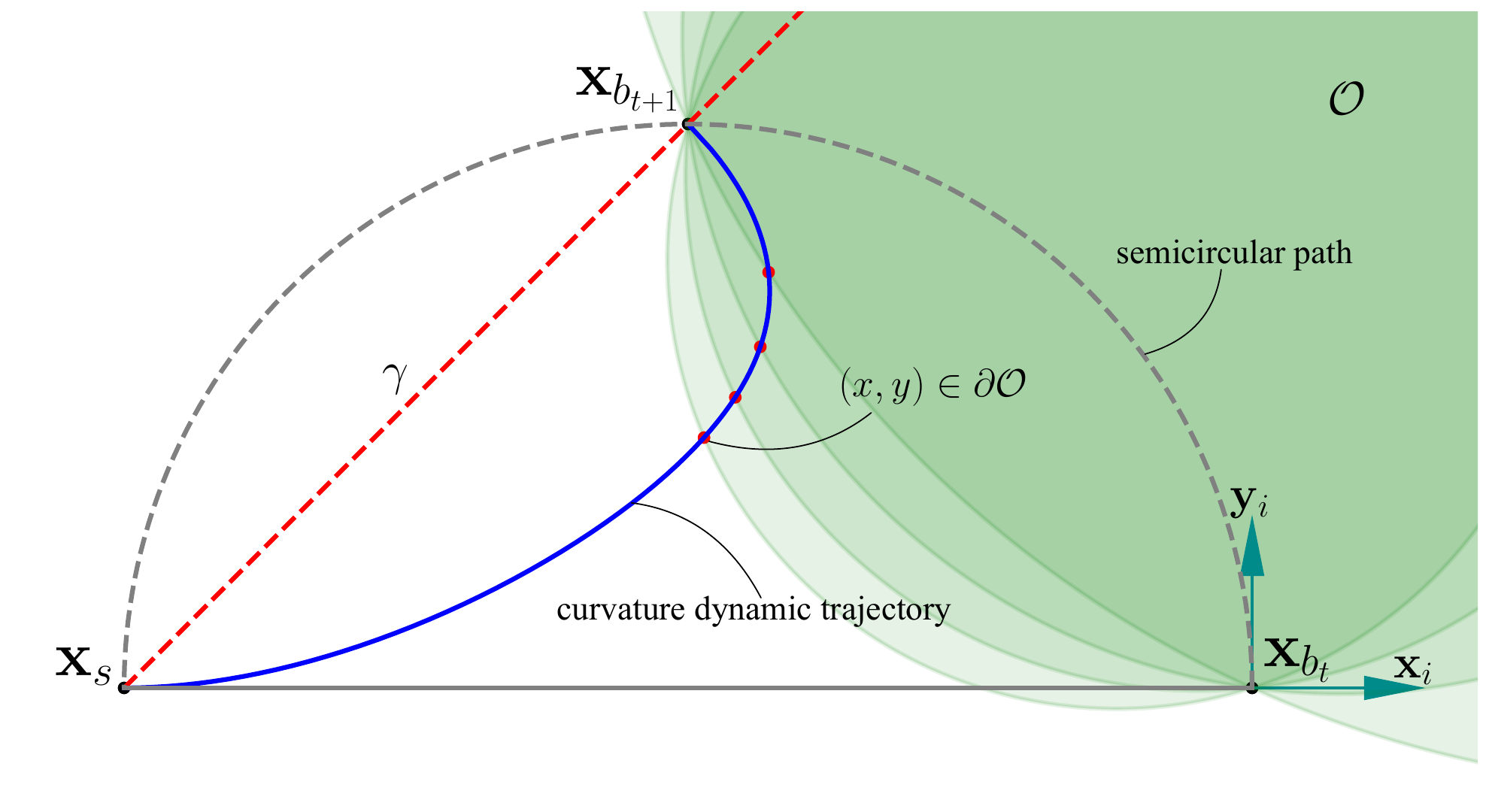}}
\caption{Curvature dynamic searching process. The decision boundary $\dout$ between two iterations is approximated by circles with different curvature. The dashed line indicates the original semicircular path, the red dots are the points $(x,y)$ with minimum $\ell_2$-norm perturbation on the estimated boundary. The solid line is the trajectory of these closest points on the circles, which is the curvature dynamic trajectory.}
\label{fig2}
\end{figure}

We propose a curvature dynamic trajectory to conduct DCE in a query-efficient way. As shown in Figure \ref{fig2}, after the semicircular path has searched another adversarial example $\x_{b_{t+1}}$ on the plane spanned by $\x_{b_t} - \x_s $ and the estimated normal vector $\hat{\bm \eta}_t$, we assume that the local 2-D decision boundary between $\x_{b_{t+1}}$ and $\x_{b_t}$ can be approximated by circles with different curvatures. The trajectory of the points with minimum $\ell_2$ perturbation on these circles is the curvature dynamic trajectory.

We write the curvature dynamic trajectory in the form of polar coordinates for the sake of binary searching (proof in Appendix \ref{apd:first}):
\begin{equation}
\begin{aligned}
    (\tan\gamma\cos\theta - \sin\theta)r^2& - r\tan\gamma + \sin\theta = 0, \\
    \text{s.t.}\; \theta \in & [0, \gamma].
\end{aligned}
\label{eq:curvature}
\end{equation}

 $\gamma$ is the angle between $\x_{b_t} - \x_s $ and $\x_{b_t} - \x_s $ . By searching on this trajectory, the algorithm will be able to find a minimum $(\hat{\rho},\hat{\theta})\in\dout$ dynamically, with respect to different curvature of the circles. And inversely, we can also get the estimated curvature by solving:
\begin{equation}
\hat{\kappa} =  ((\frac{0.5\tan\gamma}{\tan\gamma-\tan\hat{\theta}}-1)^2+(\frac{0.5\tan\gamma\tan\hat{\theta}}{\tan\gamma-\tan\hat{\theta}})^2)^{-1/2}
\end{equation}

\begin{algorithm}
\caption{One iteration of DCE}
\label{alg:1}

\textbf{Input}: {Original image $\x_s$, boundary point $\x_{b_t}\in\dout$, estimated normal vector $\hat{\bm \eta}_t$}\\
\textbf{Output}: $\x_{b_t+1}\in\dout$, $\hat{\kappa}$ 
\begin{algorithmic}[1]
        \STATE $\x_i, \y_i = \text{Schmidt\_Orthogonalization}(\frac{\x_{b_t} - \x_s}{\| \bm x_{b_t} - \bm x_s\|_2},\hat{\bm \eta}_t)$
        
        \STATE \texttt{clean} = $\x_s$, \texttt{adv} = $\x_s + \x_i$
       
        \STATE \underline{\textbf{Semicircular search}} by Eq \ref{eq:circle} until $\|\texttt{clean}-\texttt{adv}\|_2\leq\texttt{tol}$

       \STATE \underline{\textbf{Curvature Dynamic Search}} by Eq \ref{eq:curvature} between $\x_s$ and \texttt{adv} 
        \STATE \texttt{adv} = $\x_s+\hat{r}\cos\hat{\theta}\x_i+\hat{r}\sin\hat{\theta}\y_i $
        \STATE $\hat{\kappa} =  ((\frac{0.5\tan\gamma}{\tan\gamma-\tan\hat{\theta}}-1)^2+(\frac{0.5\tan\gamma\tan\hat{\theta}}{\tan\gamma-\tan\hat{\theta}})^2)^{-1/2}$ 
        \STATE Return \texttt{adv}, $\hat{\kappa}$

        \end{algorithmic}
\end{algorithm}

\subsection{Curvature Dynamic Black-box Attack (CDBA)}

\begin{figure}[htbp]
\centerline{\includegraphics[width=0.6\textwidth]{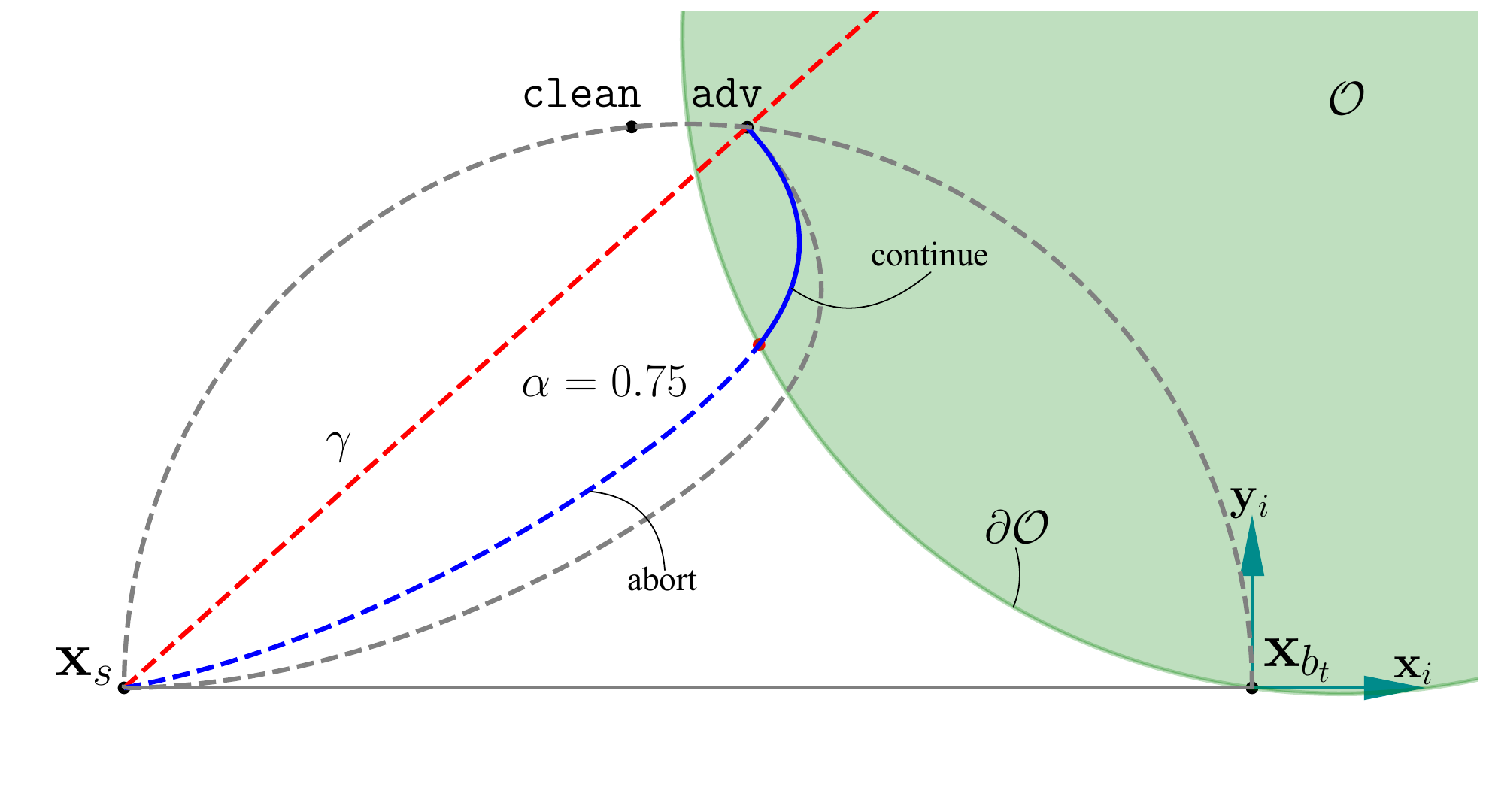}}
\caption{CDBA with abort protocol and step parameter. The gray dashed line indicates the vanilla curvature dynamic trajectory. The blue line is added step parameter $\alpha=0.75$, the dashed part is the interval to abort curvature dynamic search}
\label{fig3}
\end{figure}

Since the curvature dynamic search poses an extra query cost, we propose an abort protocol to control the trade-off between this extra query cost and the $\ell_2$-norm perturbation of the attack. Abort protocol can stop the curvature dynamic search when the curvature is low. To compensate for the error caused by this protocol and avoid reaching the local minimum, we also proposed a step parameter to control the step length in curvature dynamic searching process.

\begin{algorithm}
\caption{One iteration of CDBA}
\label{alg:2}

\textbf{Input}: {Original image $\x_s$, boundary point $\x_{b_t}\in\dout$, estimated normal vector $\hat{\bm \eta}_t$}\\
\textbf{Output}: $\x_{b_t+1}\in\dout$
\begin{algorithmic}[1]
        \STATE $\x_i, \y_i = \text{Schmidt\_Orthogonalization}(\frac{\x_{b_t} - \x_s}{\| \bm x_{b_t} - \bm x_s\|_2},\hat{\bm \eta}_t)$
        
        \STATE \texttt{clean} = $\x_s$, \texttt{adv} = $\x_s + \x_i$
       
        \STATE \underline{\textbf{Semicircular search}} by Eq \ref{eq:circle} until $\|\texttt{clean}-\texttt{adv}\|_2\leq1000\cdot\texttt{tol}$  
        \STATE $\gamma = \arctan(\frac{\langle \texttt{adv} - \x_s,\y_i \rangle}{\langle\texttt{adv} - \x_s,\x_i\rangle})$
        \STATE Solve $\theta$ for $r=\|\texttt{clean} - \x_s\|_2$ in Eq \ref{eq:curvature}
        \STATE $\theta = \gamma - \alpha*(\gamma- \theta)$
        
        \STATE \texttt{tmp} = $\x_s+r\cos\theta\x_i+r\sin\theta\y_i $
        \IF{ $\phi(\texttt{tmp}) = 1 $ }
        \STATE \underline{\textbf{Curvature Dynamic Search}} by Eq \ref{eq:curvature} with step parameter between $\x_s$ and \texttt{tmp} 
       
        \STATE Return \texttt{adv}
            \ELSE
          
            \STATE \underline{\textbf{Semicircular search}} by Eq \ref{eq:circle} between \texttt{clean} and \texttt{adv} 
            
            \STATE Return \texttt{adv}
        
        \ENDIF
        
        \end{algorithmic}
\end{algorithm}

\subsubsection{Abort protocol}
At the initial semicircular search, the search will not reach maximum tolerance in the first place. Instead, it will stop when the norm between clean example \texttt{clean} and adversarial example \texttt{adv} is smaller than 1000 times the final tolerance as shown in Figure \ref{fig3}. Then, we use \texttt{adv} to determine the tangent $\gamma$, and we check the corresponding point \texttt{tmp} with the same norm as \texttt{clean} on the trajectory, to see whether the curvature dynamic search will lead to a better result.   

If \texttt{tmp} is adversarial, then we start the curvature dynamic search between $\x_s$ and \texttt{tmp}. If not, the search will continue on the original semicircular path until it reaches the final tolerance.

\subsubsection{Step parameter}
We also found that searching along the curvature dynamic trajectory sometimes leads to a local minimum, making the descent on later iterations less ideal. Consequently, we proposed a step parameter $\alpha$ to control the step length. In the curvature dynamic searching process, instead of query at the point$(\rho^\star,\theta^\star)$, we query at the point $(\rho^\star,\gamma - \alpha*(\gamma- \theta^\star))$. By setting $\alpha \in [0,1]$, the curvature dynamic trajectory is flattened as is shown in Figure \ref{fig3}. It should be noted that when $\alpha=0$, the algorithm will be reduced to CGBA-H which uses a binary line search.

\section{Experiments}

\subsection{Experimental setting}

\subsubsection{DCE}

For $\ell_p$ robust classifiers, we selected seven WideResNet-28-10 classifiers on CIFAR-10 including three $\ell_2$ robust ones \cite{8954314,wang2023better,rebuffi2021fixingdataaugmentationimprove} and four $\ell_\infty$ robust ones \cite{pmlr-v97-hendrycks19a,NEURIPS2020_e3a72c79,zhang2021_GAIRAT,cui2021learnable}. We also selected three $\ell_\infty$ robust ResNet-50 models \cite{DBLP:conf/iclr/WongRK20,robustness,NEURIPS2020_24357dd0} on ImageNet. These robust models and the standard models are available from Robustbench library \cite{croce2021robustbench} and demonstrate different adversarial robustness. The clean accuracy and robust accuracy of these classifiers are available in Appendix \ref{apd:3}. 100 pairs of images randomly drawn from CIFAR-10 and ILSVRC2012’s validation set are used for DCE in $\ell_p$ robustness analysis.

For certified robust classifiers, we chose seven classifiers including four ResNet-101 and three ResNet-50 classifiers from \cite{pmlr-v97-cohen19c}. It used randomized smoothing, a voting strategy of samples near source image with Gaussian noise, to improve certified robustness. 50 pairs of images are used for DCE in certified robustness analysis.

For common corruption robust classifiers, three ResNet-50 robust classifiers \cite{DBLP:journals/corr/abs-1811-12231,DBLP:journals/corr/abs-2202-01263,hendrycks2021many} on ImageNet are selected. 50 pairs of images are used for DCE in common corruption robustness analysis. The classifiers and their respective robust accuracy are also available in Robustbench library \cite{croce2021robustbench}.

Considering the difference in converging speed, DCE is performed for 20 iterations on standard CIFAR-10 classifier and 30 for robust ones. DCE is performed for 40 iterations on standard ImageNet classifier and 60 for robust ones. Moreover, instead of taking the average $\hat\kappa$ over the entire iteration, the results are first divided into different bins according to their $\ell_2$-norm, curvature over 1000 is dropped because of the ill estimation around local minima. We take the average of $\hat\kappa$ for every bin in the interval we are interested in. The $\ell_2$-norm bins used for CIFAR-10 and ImageNet classifiers are \texttt{linspace(1,6,6)} and \texttt{linspace(30,80,6)} in adversarial and common corruption robustness analysis. We also take the logarithm of $\hat\kappa$ for better visualization.  

The ablation study in Appendix \ref{apd:second} includes limitation of normal vector estimation, effect of normal vector estimation and sensitivity of curvature dynamic trajectory. Moreover, we evaluated the curvature difference between targeted and non-targeted attack decision boundary in standard CIFAR-10 and ImageNet classifiers. 50 pairs of images are used for curvature analysis in different attacks.

\subsubsection{CDBA}
We evaluate CDBA using standard ResNet-50 ImageNet classifier in RobustBench \cite{croce2021robustbench}, since targeted ImageNet decision boundary has the highest curvature. We run CDBA targeted with abort protocol and step parameter $\alpha=0.75$ for 15 iterations on 100 pairs of images from ILSVRC2012’s validation set. 

For baselines, we choose CGBA and CGBA-H \cite{reza2023cgba}, the state-of-the-art non-targeted and targeted attack, which is also what DCE is based on. All hyperparameters including query number $N_0$, final tolerance, DCT factors in all experiments except ablation study are set the same as CGBA and CGBA-H for fair comparison. Note that normalization is dropped to keep aligned with the Robustbench setting.

\subsection{Experimental results}

\subsubsection{Decision boundary curvature and adversarial robustness}

\begin{table*}[ht]
\small
\centering
\renewcommand{\arraystretch}{0.8}
\begin{tabular}{|c|c|c|c|rrrrr|}

\toprule
\textbf{Datasets} & \textbf{Structure} & \multicolumn{1}{c|}{\textbf{Metric}} & \textbf{Classifier} & \multicolumn{5}{c|}{\textbf{$\overline{log(\hat\kappa)}$}} \\
\midrule
\midrule
\multirow{13}{*}{\rotatebox[origin=c]{90}{CIFAR-10}} 
& \multirow{8}{*}{WideResNet} 

& - 
& \textbf{Standard} & \textbf{-0.63} & \textbf{0.18} & \textbf{1.17} & \textbf{1.58} & \textbf{1.43}\\

\cline{3-9}

& &  \multirow{3}{*}{$\ell_2$} 

& ~\cite{8954314} & -1.75 & -1.64 & -0.93 & -0.97 & -1.41 \\
& & & ~\cite{rebuffi2021fixingdataaugmentationimprove} & \textbf{\textcolor{red}{-3.32}} & \textbf{\textcolor{red}{-3.41}} & \textbf{\textcolor{red}{-3.18}} & \textbf{\textcolor{red}{-2.78}} & \textbf{\textcolor{red}{-2.39}} \\
& & & ~\cite{wang2023better} & -3.12 & -3.33 & -2.65 & -2.34 & -1.37 \\
\cline{3-9}

& & \multirow{4}{*}{$\ell_\infty$}
& \cite{pmlr-v97-hendrycks19a} & -1.90 & -1.82 & -1.10 & -1.01 & -0.74 \\
& & & \cite{NEURIPS2020_e3a72c79} & -1.64 & -1.48 & -1.59 & -1.46 & -1.34 \\
& & & \cite{zhang2021_GAIRAT} & -1.70 & -1.55 & -1.64 & -1.37 & -1.20 \\
& & & \cite{cui2021learnable} & \textbf{\textcolor{red}{-3.03}} & \textbf{\textcolor{red}{-2.57}} & \textbf{\textcolor{red}{-2.10}} & \textbf{\textcolor{red}{-1.74}} & \textbf{\textcolor{red}{-1.53}} \\
\cline{2-9}
& \multirow{5}{*}{ResNet-101} 
& - & \textbf{Standard} & \textbf{-0.26} & \textbf{0.44} & \textbf{0.75} & \textbf{1.08} & \textbf{1.54} \\
\cline{3-9}
& & \multirow{4}{*}{Certified}  & $\sigma=0.12$ & -0.72 & -0.26 & -0.32 & 0.03 & 0.65 \\
& & & $\sigma=0.25$ & -0.65 & -0.73 & -0.49 & -0.15 & -0.96 \\
& & & $\sigma=0.50$ & -1.47 & \textbf{\textcolor{red}{-1.17}} & -0.92 & -1.36 & \textbf{\textcolor{red}{-1.35}} \\
& & & $\sigma=1.00$ & \textbf{\textcolor{red}{-2.32}} & -1.10 & \textbf{\textcolor{red}{-1.18}} & \textbf{\textcolor{red}{-1.70}} & -0.59 \\
\midrule
\multirow{7}{*}{\rotatebox[origin=c]{90}{ImageNet}} 
& \multirow{7}{*}{ResNet-50} 
& - & \textbf{Standard} & \textbf{3.24} & \textbf{3.34} & \textbf{3.39} & \textbf{3.42} & \textbf{3.44} \\
\cline{3-9}
& & \multirow{3}{*}{$\ell_\infty$} 
& ~\cite{DBLP:conf/iclr/WongRK20} & 0.59 & 1.24 & 1.93 & 1.93 & 2.14 \\
& & & ~\cite{robustness} & 0.80 & 1.60 & 1.79 & 2.36 & 2.36 \\
& & & ~\cite{NEURIPS2020_24357dd0} & \textbf{\textcolor{red}{0.17}} & \textbf{\textcolor{red}{1.07}} & \textbf{\textcolor{red}{1.45}} & \textbf{\textcolor{red}{1.91}} & \textbf{\textcolor{red}{2.10}} \\
\cline{3-9}
& & \multirow{3}{*}{Certified} & $\sigma=0.25$ & 0.91 & 0.77 & 1.12 & 1.52 & 1.70 \\
& & & $\sigma=0.50$ & 0.34 & \textbf{\textcolor{red}{0.44}} & 0.73 & 1.33 & 1.04 \\
& & & $\sigma=1.00$ & \textbf{\textcolor{red}{0.20}} & 0.48 & \textbf{\textcolor{red}{0.19}} & \textbf{\textcolor{red}{0.01}} & \textbf{\textcolor{red}{-0.07}} \\
\bottomrule
\end{tabular}
\caption{Decision boundary curvature in adversarial robust image classifiers. The greatest curvature of every basic classifier structure are marked in bold, the least curvature of every robustness metric are marked in red. Classifiers are listed in ascending order in terms of their robustness.}
\label{table:1}
\end{table*}

Table \ref{table:1} shows the average targeted decision boundary curvature $\overline{log(\hat\kappa)}$ versus different $\ell_2$-norm bins in CIFAR-10 and ImageNet classifiers. Note that the classifiers are listed in ascending order in terms of their robustness.
We observe that for all classifier structures on all datasets, the decision boundary curvature in standard classifiers is significantly higher than that in robust classifiers. Moreover, it is also apparent that the curvature in robust classifiers is associated with the exact robustness as measured by the robustness accuracy. The higher curvature of decision boundary, the more vulnerable the classifier will be to adversarial attacks.

This is especially true when the perturbation is small enough when it is closer to the perturbation of adversarial examples used to train and test these models. This might be intuitive in $\ell_2$ robust models. But such finding also holds in $\ell_\infty$ robust models, which means that adversarial robustness is generally able to generate flatter local boundaries regardless of the perturbation distribution. 

\subsubsection{Decision boundary curvature and common corruption robustness}
\begin{table}[htbp]
\small
\centering
\renewcommand{\arraystretch}{0.8}

\begin{tabular}{crrrrr}

\toprule 
\multicolumn{1}{l}{Classifier} & \multicolumn{5}{c}{$\overline{log(\hat\kappa)}$}  \\

\midrule

Standard & \textbf{3.24} & \textbf{3.34} & \textbf{3.39} & \textbf{3.42} & \textbf{3.44} \\ 
\cite{DBLP:journals/corr/abs-1811-12231} & 2.75	&2.99	&3.08	&3.05	&3.20\\ 

\cite{DBLP:journals/corr/abs-2202-01263} & \textcolor{red}{2.71}	&2.85 &3.03	&3.17	&3.29 \\
\cite{hendrycks2021many} & 2.87	&\textcolor{red}{2.73}	&\textcolor{red}{2.68}	&\textcolor{red}{2.77}	&\textcolor{red}{2.85} \\ 
\bottomrule
\end{tabular}
\caption{Decision boundary curvature in common corruption robust models}
\label{tab:3}
\end{table}

In common corruption robust classifiers, however, things are a little bit different. As is shown in Table \ref{tab:3}, although the decision boundary curvature is still slightly lower in robust classifiers, they don't seem to be strongly related to their exact robustness. This might be because the perturbation of common corruptions is generally higher than adversarial perturbations, which doesn't necessarily result in flattened local boundary as shown in adversarial training.

\subsubsection{CDBA}
We run CGBA, CGBA-H, CDBA $(\alpha=1,0.75)$ separately for 15 iterations of targeted attack on standard ResNet-50 classifier to show the average $\ell_2$-norm versus query number results in different attack methods. The least $\ell_2$ perturbation is marked in bold.

\begin{table}[htbp]
\small
\centering
\renewcommand{\arraystretch}{0.8}
\begin{tabular}{lcccc} 
\toprule 
Query & 250 & 500 & 750 & 1000  \\ 
\midrule 
CGBA & 96.93 & 94.11 &91.63 & 89.05 \\ 
CGBA-H & 82.78 & 74.13 & 68.60 & 63.91 \\ 
CDBA$(\alpha=1)$ & \textbf{81.72} & 73.86 & 68.37 & 64.23  \\
CDBA$(\alpha=0.75)$ & 82.25 & \textbf{73.85} & \textbf{67.69} & \textbf{62.78}  \\

\bottomrule
\end{tabular}
\caption{mean $\ell_2$-norm distortion in targeted attack on ImageNet}
\label{tab:5}
\end{table}

From Table \ref{tab:5} we observe that CDBA($\alpha=0.75$) achieves the least $\ell_2$-norm perturbation under a limited query budget 1000. Moreover, standard CDBA outperforms other attacks during the initial descent, with relatively high curvature of decision boundary. In later iterations, as perturbation decreases, the step parameter added successfully compensates for the error caused in the previous abort protocol and prevents the algorithm from reaching the local minima. The effectiveness of step parameter within each iteration will be further discussed in ablation study.

\section{Conclusion}
In this paper, we proposed Dynamic Curvature Estimation, which performs query-efficient estimation of curvature during the process of black-box attacks. By conducting DCE on standard and robust image classifiers, we revealed a connection between decision boundary curvature and adversarial robustness. We also designed an ablation study on the proposed curvature dynamic trajectory to prove its effectiveness. Furthermore, using curvature dynamic search, we also crafted a new decision-based adversarial attack CDBA, which uses the curvature information to improve the attack under a limited query budget.

\bibliography{pmlr-sample}

\appendix

\section{Proof for curvature dynamic trajectory}\label{apd:first}

We assume that the local 2-D decision boundary between $\x_{b_{t+1}}$ and $\x_{b_t}$ can be approximated by circles with different curvatures. 

For these circles to pass $\x_{b_t}$, the equation of the set of circles can be written as:
\begin{equation}
{(x-x_0)}^2 + {(y-y_0)}^2 = {(x_0-1)}^2 + y_0^2
\label{eq}
\end{equation}

where $(x_0,y_0)$ is the center of circle.

Denote the angle between $\x_{b_t} - \x_s $ and $\x_{b_t} - \x_s $ as $\gamma$. For the circles to pass both $\x_{b_t}$ and $\x_{b_{t+1}}$, the center should be on the perpendicular bisector of them, which is:
\begin{equation}
y_0=\tan\gamma(x_0-0.5)
\label{11}
\end{equation}

Finally, the closest points $(x,y)$ to $\x_s$ on the circles should also satisfy that:
\begin{equation}
y=\frac{y_0}{x_0}x
\label{12}
\end{equation}

By solving Eq\ref{11} and Eq\ref{12}, we can get:

\begin{equation}
\begin{aligned}
    x_0 &= \frac{0.5\tan\gamma\cdot x}{\tan\gamma\cdot x-y},\\
    y_0 &= \frac{0.5\tan\gamma\cdot y}{\tan\gamma\cdot x-y}.
\end{aligned}
\label{14}
\end{equation}
By inserting Eq\ref{14} into Eq \ref{eq}, we can get the trajectory of the points with minimum $\ell_2$ perturbation on these circles, which is the curvature dynamic trajectory:
\begin{equation}
\begin{aligned}
    (\tan\gamma\cos\theta - \sin\theta)r^2& - r\tan\gamma + \sin\theta = 0, \\
    \text{s.t.}\; \theta \in & [0, \gamma].
\end{aligned}
\end{equation}

\section{Ablation study}\label{apd:second}

\subsection{Limitation of normal vector estimation}
As mentioned before, the estimated normal vector on the entire space(even after the dimension reduction via DCT) differs greatly from the normal vector in the restricted plane because the query number is far less than the dimension. Here we propose an experiment with normal vector estimation on the restricted plane immediately after the initial full space normal vector estimation in each iteration. We first generate a set of random vectors $\{\x_k\}_{k=1}^{K}$ and project them to $\{\x_k^{\text{proj}}\}_{k=1}^{K}$ on the plane using: 
\begin{equation}
    \x_k^{\text{proj}} = \langle\x_k,\x_i\rangle\x_i + \langle\x_k,\y_i\rangle\y_i
\end{equation}

where $\x_i$ and $\y_i$ are the xy basis of the plane. Then $\{\x_k^{\text{proj}}\}_{k=1}^{K}$ is normalized to $\bm \delta_k^{\text{proj}} \sim \mathcal N(0, \mathbf{\sigma^2})$. We can obtain the normal vector on the plane by:
\begin{equation} 
    \hat{\bm \eta}_t^{\text{proj}} = \frac{\sum_{k=1}^{K} \phi(\x_{b_t} + \bm\delta_i^{\text{proj}})\bm \delta_i^{\text{proj}}}{\|\sum_{k=1}^{K} \phi(\x_{b_t} + \bm \delta_i^{\text{proj}})\bm \delta_i^{\text{proj}}\|_2} .
\end{equation}
where $\phi(\cdot)$ is the hard label indicator function. The error angle between $\hat{\bm \eta}_t^{\text{proj}}$ and $\hat{\bm \eta}_t$ can be obtained by $\psi_t=\arccos\frac{\langle\hat{\bm \eta}_t,\hat{\bm \eta}_t^{\text{proj}}\rangle}{\|\hat{\bm \eta}_t\|_2\cdot\|\hat{\bm \eta}_t^{\text{proj}}\|_2}$

Specifically, we choose $\sigma=0.0002$, which is the same as full space normal vector estimation in CGBA. Four initial query number $N_0$ are selected for normal vector estimation with dimension reduction factor $f=4$ and $K=100$ for normal vector estimation within the restricted plane. The experiment is conducted for 20 iterations on 100 pairs of random images in standard CIFAR-10 classifier.

\begin{table}[htbp]
\centering
\renewcommand{\arraystretch}{0.8}

\begin{tabular}{lcccc} 
\toprule $N_0$ & 10 & 20 & 30 & 40  \\ 
\midrule 
MSE$(^\circ)$ & 48.34 & 43.28 & 40.41 & 37.69 \\ 
\bottomrule
\end{tabular}
\caption{MSE of normal vector estimation with different initial query number $N_0$}
\label{tab:error}
\end{table}

Table \ref{tab:error} shows the MSE of the normal vector estimation  measured by $\psi_t$ between $\hat{\bm \eta}_t^{\text{proj}}$ and $\hat{\bm \eta}_t$ in degree. Since the error is quite large, it is barely possible to utilize the geometric property of normal vectors in the restricted plane. This is why we propose the circular interpolation of decision boundary in DCE regardless of the normal vector. However, in DCE we still follow the normal vector estimation setting in CGBA since it successfully minimizes the $\ell_2$ perturbation in a query-efficient way. This might be because although there is estimation error, the vector is still potentially pointing towards the vast convex adversarial region. 

\subsection{Effect of normal vector estimation in curvature estimation}
Normal vector estimation can still affect curvature estimation in a way that it determines the direction from the whole space to construct the restricted plane. To reveal its effect, we compare 300 random estimated curvature with $\ell_2$-norm $\in[1,2)$ in the previous experiment. Similarly, estimated curvature over 1000 is dropped empirically because of bad estimation around local minima.

\begin{table}[htbp]
\centering
\renewcommand{\arraystretch}{0.8}

\begin{tabular}{lcccc} 
\toprule $N_0$ & 10 & 20 & 30 & 40  \\ 
\midrule 
${log(\hat\kappa)}$ & $1.79_{\pm1.11}$& $1.64_{\pm1.00}$ & $1.51_{\pm1.07}$ & $1.49_{\pm1.11}$ \\ 
\bottomrule
\end{tabular}
\caption{${log(\hat\kappa)}$ with different initial query number $N_0$}
\label{tab:aba1}
\end{table}

Table \ref{tab:aba1} shows the mean and standard deviation of the logarithm of the estimated curvature in four $N_0$. We observe a modest drop in curvature as query number increases. An ANOVA test of four sets is conducted with a result of $F=4.88$ and $p=0.002$, indicating there is a significant shift in the mean curvature using different initial query numbers. Normal vector estimation with higher query number can not only result in reduced error, but also points towards flatter adversarial region.

\subsection{Sensitivity of curvature dynamic trajectory}
We compare the result of CGBA-H and three DCE steps $(\alpha=0.5, 1, 1.5)$ without abort protocol in each iteration to reveal the effectiveness of curvature dynamic trajectory. The experiment is conducted for 10 iterations on 100 pairs of random images in standard CIFAR-10 classifier. We record their mean $\ell_2$-norm decrease percentage in each iteration.  

\begin{figure}
    \centering
    \includegraphics[width=0.5\linewidth]{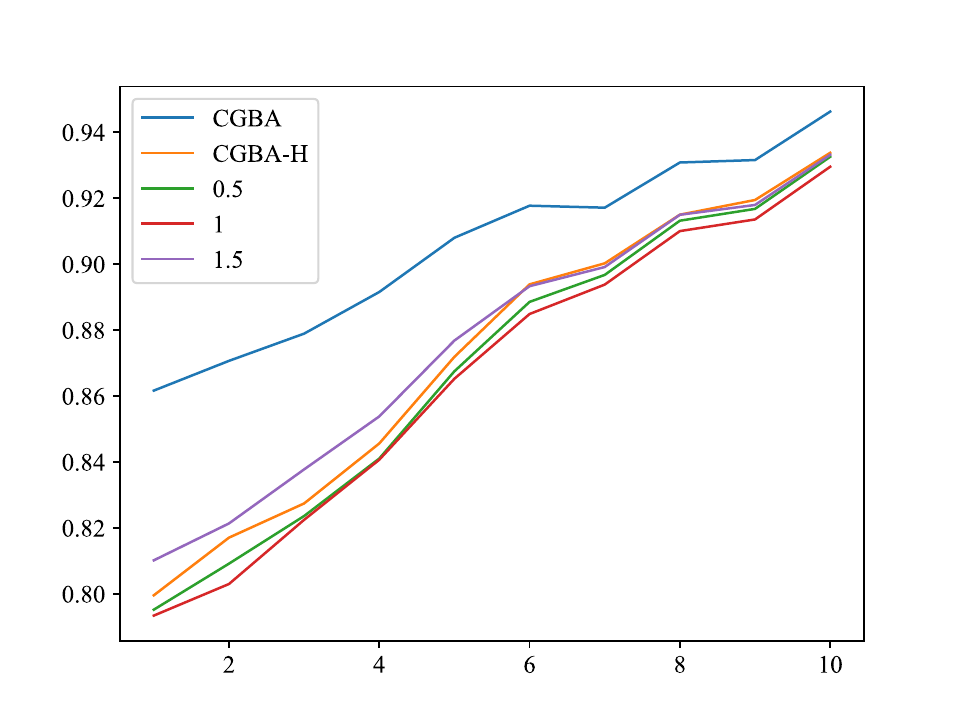}
    \caption{$\|\x_{b_{t+1}}\|_2/|\x_{b_{t}}\|_2 $ in first 10 iterations}
    \label{fig5}
\end{figure}

As shown in Figure \ref{fig5}, DCE with $\alpha=1$ achieves the best performance over other step parameters and CGBA-H per iteration, indicating that the vanilla DCE can best locate the minima on the restricted plane. It also means that the local circular assumption of decision boundary is effective in constructing the curvature dynamic trajectory. 

\subsection{Targeted and Non-targeted decision boundary curvature}

\begin{table}[htbp]
\centering
\renewcommand{\arraystretch}{0.8}

\begin{tabular}{lrrrr}


\toprule 
\multicolumn{1}{l}{Dataset} & \multicolumn{4}{c}{$\ell_2$-norm bins }  \\

\toprule
CIFAR-10 & [0,1) & [1,2) & [2,3) & [3,4)   \\ 
\midrule 
Targeted & \textbf{-1.66} & \textbf{-0.63} & \textbf{0.18} & \textbf{1.17}  \\ 

Non-targeted & \textcolor{red}{-2.38} & \textcolor{red}{-1.77}  &  \textcolor{red}{-1.19} &  \textcolor{red}{-0.82}  \\
\toprule
ImageNet & [0,10) & [10,20) & [20,30) & [30,40)  \\ 
\midrule 
Targeted & \textbf{2.23} & \textbf{2.43} & \textbf{2.93} & \textbf{3.24}  \\ 

Non-targeted & \textcolor{red}{-0.67} & \textcolor{red}{0.12}  &  \textcolor{red}{-0.37} &  \textcolor{red}{-0.89}  \\
\bottomrule
\end{tabular}
\caption{$\overline{log(\hat\kappa)}$ versus $\ell_2$-norm for different attacks}
\label{tab:4}
\end{table}
Table \ref{tab:4} shows the curvature difference in targeted and non-targeted decision boundaries. The $\ell_2$-norm bins used here are slightly different since the attack converges faster in non-targeted scenario. It is clear that the decision boundary in targeted attacks is higher than that in non-targeted attacks, which is in accordance with previous research. 

Although the non-targeted decision boundary curvature is actually lower, most attacks including CGBA converge faster in non-targeted attacks. This means that the estimated curvature cannot be simply interpreted as an indicator of converging speed or attack success rate and is thus associated with robustness. Higher curvature is not a direct result of better attack results.

\section{Adversarial robustness of models}\label{apd:3}
\begin{table}[htbp]
\centering
\renewcommand{\arraystretch}{0.8}

\begin{tabular} {c|c|c|c|}
    \toprule
     \multicolumn{1}{c|}{} &
      \multicolumn{1}{c|}{Classifier}  & \multicolumn{1}{c|}{Clean acc.} & \multicolumn{1}{c|}{Robust acc.}\\
      
    \midrule 
    \multirow{4}{*}{\rotatebox[origin=c]{90}{CIFAR($\ell_2$)}} 
     
    & Standard  & 94.78\% & 0\\ 
    &~\cite{8954314}  & 89.05\% & 66.44\% \\\
   
    & ~\cite{rebuffi2021fixingdataaugmentationimprove}  & 91.79\% & 78.80\% \\ 
    & ~\cite{wang2023better}  & 95.16\% & 83.68\% \\
   
    \midrule 
    \multirow{4}{*}{\rotatebox[origin=c]{90}{CIFAR($\ell_\infty$)}} 
     
    & Standard  & 94.78\% & 0\\ 
    &~\cite{pmlr-v97-hendrycks19a}  & 87.11\% & 54.92\% \\\
    &~\cite{NEURIPS2020_e3a72c79} & 88.98\% & 57.14\% \\\
   
    & ~\cite{zhang2021_GAIRAT} & 89.36\% & 59.64\% \\ 
    & ~\cite{cui2021learnable}  & 92.16\% & 67.73\% \\
    \midrule
    \multirow{4}{*}{\rotatebox[origin=c]{90}{ImageNet}} & Standard & 76.52\% & 0\\
    & ~\cite{DBLP:conf/iclr/WongRK20} & 55.62\% & 26.24\%\\

    & ~\cite{robustness} & 62.56\% & 29.22\%\\ 
    & ~\cite{NEURIPS2020_24357dd0}  & 64.02\% & 34.96\%\\
     \bottomrule 
\end{tabular}   
\caption{Clean and robust accuracy of $\ell_p$ robust classifiers on two datasets}
\label{tab:2} 
\end{table}

\end{document}